\documentclass[sigconf]{acmart}

\usepackage{amsmath,amssymb,amsfonts}
\usepackage{algorithmic}
\usepackage{subfigure}
\usepackage{textcomp}
\usepackage{xcolor}

\usepackage{array}
\usepackage{multirow}
\usepackage{mathtools}
\usepackage{amsthm}

\usepackage{dcolumn}
\newcolumntype{d}[1]{D{.}{.}{#1}}

\newtheorem{definition}{Definition}

\newtheorem{property}{Property}

\DeclarePairedDelimiterX{\infdivx}[2]{(}{)}{%
	#1\;\delimsize\|\;#2%
}
\DeclarePairedDelimiter{\norm}{\lVert}{\rVert}

\AtBeginDocument{%
  \providecommand\BibTeX{{%
    \normalfont B\kern-0.5em{\scshape i\kern-0.25em b}\kern-0.8em\TeX}}}

\setcopyright{none}

\begin{document}

\title{An Improved Historical Embedding without Alignment}


\author{Xiaofei Xu}
\affiliation{%
  \institution{Southwest University}
  \state{Chongqing Shi}
  \country{China}
  }
\email{nakamura@email.swu.edu.cn}

\author{Ke Deng}
\affiliation{%
  \institution{RMIT University}
  \state{VIC}
  \country{Australia}
  }
\email{ke.deng@rmit.edu.au}

\author{Fei Hu}
\affiliation{%
  \institution{Chongqing University of Education}
  \state{Chongqing Shi}
  \country{China}
  }
\email{etz1@163.com}

\author{Li Li}
\affiliation{%
  \institution{Southwest University}
  \state{Chongqing Shi}
  \country{China}
  }
\email{lily@swu.edu.cn}



\begin{abstract}
Many words have evolved in meaning as a result of cultural and social change. Understanding such changes is crucial for modelling language and cultural evolution. Low-dimensional embedding methods have shown promise in detecting words' meaning change by encoding them into dense vectors. However, when exploring semantic change of words over time, these methods require the alignment of word embeddings across different time periods. This process is computationally expensive, prohibitively time consuming and suffering from contextual variability. In this paper, we propose a new and scalable method for encoding words from different time periods into one dense vector space. This can greatly improve performance when it comes to identifying words that have changed in meaning over time. We evaluated our method on dataset from Google Books N-gram. Our method outperformed three other popular methods in terms of the number of words correctly identified to have changed in meaning. Additionally, we provide an intuitive visualization of the semantic evolution of some words extracted by our method.

\end{abstract}




\maketitle

\section{Introduction}
Embedding words into a low-dimensional vector space as vectors according to their co-occurrence statistics has shown promise as a method in many Natural Language Processing tasks such as next-word prediction \cite{bengio2003neural} and sentiment analysis \cite{mikolov2013linguistic}. Given a word $w$ in a corpus, the co-occurrence (also called \textit{context}) of $w$ refers to the words which appear next to $w$ within a range $L$ (e.g., if $L=2$ it includes the two words before and after $w$ in all sentences in the corpus). Based on the distributional hypothesis that word semantics are implicit in co-occurrence relationships \cite{Harris:1954}, the semantic similarity between two words can be approximated by the cosine similarity (or distance) between their word embeddings \cite{turney2010frequency} \cite{hamilton2016diachronic}. As a consequence, complex linguistic problems, such as exploring semantic change of words in discrete time periods \cite{sagi2011tracing} \cite{wijaya2011understanding} \cite{gulordava2011distributional} \cite{jatowt2014framework} can thus be tackled properly. Moreover, embedding methods have been used to detect large scale linguistic change-point \cite{kulkarni2015statistically}, quantify changes in social stereotypes \cite{garg2018word} as well as to seek out regularities in acquiring language, such as the attempt to undergo parallel change over time \cite{xu2015computational}.

SVD (Singular Value Decomposition) and SGNS (Skip-Gram of word2vec with Negative Sampling) are the typical low-dimensional embedding methods and have been extensively used in language diachronic analysis \cite{hamilton2016cultural} \cite{grayson2017exploring} \cite{levy2015improving}. Along with time, the corpus evolves and accordingly the co-occurrence statistics of words change. Using SVD and SGNS on the corpus segments in different time periods, words will be embedded into separate vector spaces, each for one corpus segment, and thus cannot be effectively compared across time \cite{hamilton2016diachronic}. To address this issue, the existing studies normally encode words first into separate vector spaces in different time periods and then align the learned word embeddings across time. The two steps can be done separately \cite{gower2004procrustes} \cite{hamilton2016diachronic} \cite{hamilton2016cultural} \cite{7511732} or concurrently \cite{Yao:2018} \cite{Rudolph:2018} \cite{Robert:2017}. Such alignment is based on the assumption that most words remain unchanged. So, the alignment objective is to minimize the overall distance of word embeddings across different vector spaces. However, none of the existing low-dimensional embeddings ensures the alignment is smooth, i.e., if a word has the more similar co-occurrence statistics at different time periods, the word embeddings at these time periods tend to be more similar; otherwise, the word embeddings tends to be more dissimilar. 

Some efforts have been made to circumvent the problems of alignment by not encoding words into low-dimensional vector spaces \cite{gulordava2011distributional} \cite{jatowt2014framework}. Among them, the Positive Point-wise Mutual Information (PPMI) \cite{turney2010frequency} outperforms a wide variety of other high-dimensional approaches \cite{bullinaria2007extracting}. PPMI naturally aligns word vectors smoothly by constructing a high-dimensional sparse matrix where each row represents a word in a vocabulary, each column represents a word in the same vocabulary, and the element value indicates whether the word of a column is in the context of the word of a row based on co-occurrence in a corpus. Although PPMI wards off alignment issues, it does not enjoy the advantages of low-dimensional embeddings such as higher efficiency and better generalization. That is, tracing PPMI will consume a lot of computing resources in high-dimensional sparse environment. It also brings bias towards the infrequent events, i.e., an infrequent word in context often brings to its corresponding word a higher chance to change \cite{levy2015improving}.

In this paper, we propose \textit{Tagged-SGNS} (TSGNS) which extends SGNS by incorporating the corpus segments in different time periods. TSGNS enjoys the high performance of low-dimensional embeddings as SGNS and the smooth alignment of vector spaces across different time periods of the high-dimensional approaches as PPMI. Also, it is worthy to mention using the scheme of TSGNS one can also extend SVD to \textit{Tagged-SVD} (TSVD) for smooth alignment. However, TSVD is less preferable due to the much higher requirement on memory and thus this paper focuses on TSGNS. To verify the effectiveness of TSGNS, we have conducted extensive experiments on Google Books N-gram dataset (105GB), MEN dataset (3000 word pairs with human labelled similarity), and a dataset from Oxford Dictionaries (412 words with human-recognised semantic shift over time). Experimental results show the unique advantage of TSGNS against the current state-of-the-art. Our contributions are summarized as follows:

\begin{itemize}
\item This study proposes the concept of smooth alignment for word embedding over time which is a desirable property in diachronic analysis but is not held using the current state-of-the-art.
\item This study proposes innovative TSGNS based on SGNS for embedding words which are smoothly aligned across different time periods by projecting them into a common low-dimensional dense vector space.
\item This study verifies the effectiveness of TSGNS on a large dataset against the current state-of-the-art in diachronic analysis.
\end{itemize}

The rest of the paper is organized as follows. Section \ref{sec:literature} reviews the related work in diachronic analysis. Section \ref{sec:method} provides the details of the proposed method and how it solves the smooth alignment problem. Section \ref{sec:test} evaluates the proposed method against the current state-of-the-art thoroughly on an 105GB Google Books N-gram dataset. Finally, the paper is concluded in Section \ref{sec:conclusion}.

\section{Related Work}\label{sec:literature}
Diachronic analysis of words has attracted many attentions recently \cite{hamilton2016cultural} \cite{hamilton2016diachronic} \cite{kulkarni2015statistically} \cite{Liao:2016} \cite{7511732} \cite{Yao:2018} \cite{Roberto:2018} \cite{Hosein:2017}. 

\subsection{Word Embedding}\label{sec:related1}
In PPMI, words are represented by constructing a high-dimensional sparse matrix $\mathbf{M}\in \mathbb{R}^{|V_w|\times|V_c|}$ where $V_w$ and $V_c$ are the word and context vocabularies respectively. In $\mathbf{M}$, each row denotes a word $w$ in the word vocabulary and each column represents word $c$ in the context vocabulary. Typically, $V_w\equiv V_c$. So, we simply use $V$ to represent $V_w$ and $V_c$ in the rest of this paper. Let PPMI($L$) be the word embeddings using PPMI based on a corpus with the range of context $L$. For example, if $L=2$, the context includes the two words before and after $w$ in all sentences in the corpus. In PPMI($L$), the value of matrix element $M_{ij}$ in $\mathbf{M}$ suggests the associated relationship between the word $w_i$ and the context word $c_j$, estimated by:
\begin{equation}\label{eq1}
\begin{aligned}
M_{ij}&=max\Big\{\log\Big(\frac{\hat p(w_i,c_j)}{\hat p(w_i)\hat p(c_j)}\Big),0\Big\} \\
&=max\Big\{\log\Big(\frac{\#(w_i,c_j)\cdot |D|}{\#(w_i)\cdot\#(c_j)}\Big),0\Big\}
\end{aligned}
\end{equation}
where $|D|$ is the total number of sentences in corpus $D$, $\hat p(*)$ and $\hat p(*,*)$ correspond to the normalized empirical probabilities of word and joint probabilities of two words respectively; $\#(*,*)$ is the number of times the word-context pair $(*,*)$ appears in the corpus and $\#(*)$ is the number of times word $(*)$ appears in the corpus. If the word-context pair $(w_i,c_j)$  is not observed in the corpus (i.e., $\#(w_i,c_j)=0$), the function $\log()$ goes to negative infinity. In order to alleviate the problem, the $max()$ function is introduced to ensure the element value finite and greater than zero \cite{bullinaria2007extracting}.

SVD word embeddings corresponds to low-dimensional approximation of PPMI word embeddings learned via singular value decomposition. It decomposes the sparse matrix $\textbf{M}$ into the product of three matrices, $\mathbf{S}=\mathbf{U}\cdot \mathbf{\Sigma}\cdot \mathbf{I}^\top$, where both $\mathbf{U}$ and $\mathbf{I}$ are orthogonal, and $\mathbf{\Sigma}$ is a diagonal matrix of singular values ordered in the descent direction. In $\mathbf{\Sigma}$, a small number of the highest singular values retain most features of words, that is, by keeping the top $d$ singular values. We can have $\mathbf{S}_d=\mathbf{U}_d\cdot \mathbf{\Sigma}_d\cdot \mathbf{I}_d^\top$ to approximates $\mathbf{M}$. So, the word embeddings $\mathbf{W}$ is approximated by
\begin{equation}\label{eq:svd}
\mathbf{W}\approx \mathbf{U}_d
\end{equation}
Compared to PPMI, SVD representations can be more robust, as the dimension reduction acts as a form of regularization \cite{hamilton2016diachronic}.

In SGNS, each word $w$ is represented by two dense and low dimensional vectors, a word vector $\vec w_i$ and context vector $\vec w_{i+j}$ \cite{mikolov2013efficient}. The structure of SGNS consists of an input layer, a hidden layer and an output layer. Training word embeddings are optimized by maximizing the average log probability as follows:
\begin{equation}\label{eq3}
\displaystyle \mathop{\arg\max}\Big\{\frac{1}{|V|}\sum_{i=1}^{|V|}\sum_{-L\leqslant j\leqslant L, j\neq 0}\log p(w_{i+j}|w_i)\Big\}
\end{equation}
where $w_i$ is the word, $w_{i+j}$ is the context word of $w_i$, and $L$ is the context range. The bigger the $L$, the more running time costed while the more accurate the prediction. $p(*|*)$ is a softmax function:
\begin{equation}\label{eq4}
\displaystyle p(w_{i+j}|w_i)=\frac{\exp(\vec w_{i+j}^\top \vec w_i)}{\Sigma_{x=1}^{|V|}\exp((\vec w_{x})^\top \vec w_i)}
\end{equation}
where $\vec w_{*}^\top \vec w_i$ is the value of node $w_{*}$ in the output layer. SNGS has the benefit of allowing incremental initialization during learning where embeddings for time $t$ are initialized with the embeddings from time $t-\Delta$ \cite{hamilton2016diachronic} \cite{Kim:2014}.

\subsection{Aligning Vector Spaces Across Time}
In PPMI, being a sparse embedding method, each column of the matrix $M$ corresponds to one word in the context vocabulary. Using Eq. (\ref{eq1}), $M_{ij}$ can be calculated in a time period by computing $\hat p(w_i,c_j)$ based on the corresponding corpus segment and keeping $\hat p(w_i)$ and $\hat p(c_j)$ on the whole corpus. Thus, the PPMI embeddings are naturally aligned. For low-dimensional embedding methods, i.e., SGNS and SVD, words at different time periods are embedded into separate vector spaces. In order to compare word vectors from different time periods, we must ensure that the vectors are aligned \cite{hamilton2016diachronic}.

In \cite{kulkarni2015statistically}, a linear transformation of words between any two time periods is found by solving a $d$-dimensional least square problem of $k$ nearest neighbor words (where $d$ is the embedding dimensions). In \cite{7511732}, a linear transformation approach between a base and target time slices is applied and computed using anchor works, i.e., the words without change of meaning in the two time slices. Orthogonal Procrustes analysis\cite{gower2004procrustes} is the prevalent way to align the learned dense embeddings \cite{hamilton2016cultural}, it imposes the transformation to be orthogonal and solves a $d$-dimensional Procrustes problem between every two adjacent time slices. It assumes that most of the words are stable (their meanings) or change little over time. Then one can align dense embeddings by optimizing
\begin{equation}\label{eq5}
\displaystyle  \vec Q=argmin(\Sigma_{i=1}^{|V_w|}\left\|\vec w_i^{(t)}\vec Q-\vec w_i^{(t+1)}\right\|_F)
\end{equation}
where word embeddings $\vec w_i^{(t)}$ and $\vec w_i^{(t+1)}$ are learned in time periods $t$ and $t+1$ respectively, and $\vec Q$ is the aligning matrix which projects dense embeddings in vector spaces in $t$ and $t+1$ into a common vector space.

In addition to enforcing pairwise alignment, Yao et al. \cite{Yao:2018} proposed finding temporal word embeddings by enforcing alignment across all time periods. In \cite{Yao:2018}, a joint optimization problem:
\begin{equation}\label{eq:yao}
\begin{split}
\displaystyle \min_{W^{(1)},\dots,W^{(T)}}\frac{1}{2}&\sum_{t=1}^{T}\norm{\mathbf{M}^{(t)}-W^{(t)}(W^{(t)})^\top}_F^2+\\
&\frac{\lambda}{2}\sum_{i=1}^T\norm{W^{(t)}}_F^2+\frac{\tau}{2}\sum_{i=2}^T\norm{W^{(t-1)}-W^{(t)}}_F^2
\end{split}
\end{equation}
where $\mathbf{M}^{(t)}$ is PPMI$(t,L)$ (PPMI$(L)$ is represented as PPMI$(t,L)$ if we consider the corpus segment in time period $t$ only), and $\lambda,\tau>0$. Here the penalty term $\norm{W^{(t)}}_F^2$ enforces the low-rank data-fidelity. The key smoothing term $\norm{W^{(t-1)}-W^{(t)}}_F^2$ encourages the word embeddings to be aligned. The parameter $\tau$ controls how fast we allow the embeddings to change; $\tau=0$ enforces no alignment, and picking $\tau\rightarrow \infty$ converges to a static embedding with $W^{(1)}=W^{(2)}=\dots=W^{(T)}$. As indicated by authors, a key challenge is that for large vocabulary $V$ and many time periods $T$ one cannot fit all PPMI matrices in memory since they are sparse matrices. 

Recently, the dynamic word embeddings have been investigated \cite{Rudolph:2018} \cite{Robert:2017}. In \cite{Rudolph:2018}, the solution is based on Bernoulli embedding across time. It is characterised by regularizing the Bernoulli embedding with placing priors on the embedding. In \cite{Robert:2017}, it generalizes the skip-gram model to a dynamic setup where word and context embeddings evolve in time according to a diffusion process. 

All above alignment methods discussed above cannot guarantee the alignment is smooth. Due to the assumption that the meaning of most words did not shift over time, they try to minimize the overall distance between embeddings of same words across different time periods. It allows to sacrifice (i.e., distort) the distance for some words to achieve the better overall distance for all words. In other words,  a word has the more similar co-occurrence statistics at different time periods, the word embeddings at these time periods may be more similar or dissimilar.    


\subsection{Others}
The deep structured learning has been explored in word representation. In \cite{peters2018deep}, the deep bidirectional language models (e.g., LSTM) is applied and a feature-based deep contextualized word representation method known as ELMo (Embeddings from Language Models) has been proposed. In \cite{devlin2018bert}, a fine-tuning based deep contextualized word representation method known as BERT (Bidirectional Encoder Representations from Transformers) has been proposed. These methods aim to provide context-sensitive word embeddings. i.e., a word may have different representations in different sentences since other words in the sentence make the word semantically unique. This is a different problem from ours in this study.

\section{Methodology}\label{sec:method}
In practice, a word $w$ typically does not have exactly same co-occurrence statistics in two corpus segments over time. Let the embeddings of word $w$ be $\vec{w}^{(t)}$ and $\vec{w}^{(t+1)}$ based on the corpus segments in two time periods $t$ and $t+1$ respectively. The similarity between $\vec{w}^{(t)}$ and $\vec{w}^{(t+1)}$ is gauged using cosine similarity. The more similar $\vec{w}^{(t)}$ and $\vec{w}^{(t+1)}$ are, the less the difference between $w$'s co-occurrence statistics over time is. Formally, we propose the following concept.

\begin{definition}[Smooth Alignment]
Given a corpus, let $D^{(t)}$ and $D^{(t+1)}$ be the segments of a corpus $D$ in two time periods $t$ and $t+1$ respectively. Given any word $w$ in a vocabulary $V$, it is represented as $w^{(t)}$ in time period $t$ and is represented as $w^{(t+1)}$ in time period $t+1$. The word embedding of $w^{(t)}$ and $w^{(t+1)}$, denoted as $\vec{w}^{(t)}$ and $\vec{w}^{(t+1)}$, are based on $D^{(t)}$ and $D^{(t+1)}$ respectively. The word embeddings are \textit{smoothly aligned} if the following condition is satisfied, that is, $\vec{w}^{(t)}$ and $\vec{w}^{(t+1)}$ tend to be more similar if more words co-occur in the context of $w$ in $D^{(t)}$ and $D^{(t+1)}$; less similar otherwise.
\end{definition}


The smooth alignment is desirable. It ensures that the distance between word embeddings is only because of the difference between co-occurrence statistics of the word in different corpus segments over time, rather than because of the improper vector space alignment.

\begin{figure}[t]
	\centering
	\includegraphics[width=0.5\textwidth]{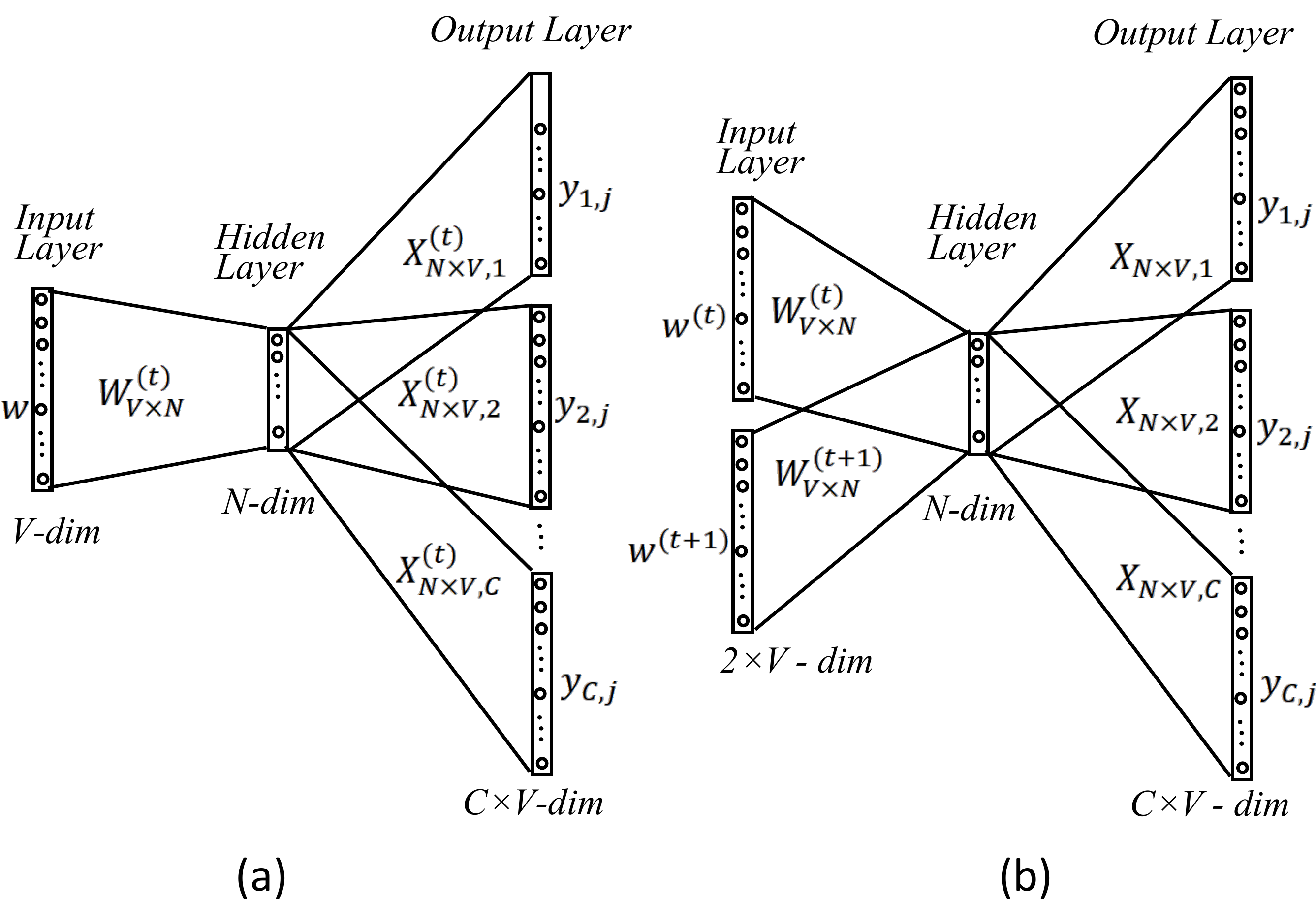}
	\caption{(a) The skip-gram model. (b) The tagged skip-gram model.}
	\label{fig:TSGNS}
\end{figure}

\subsection{Tagged-SGNS (TSGNS)}\label{sec:tsgns}
TSGNS is based on SGNS. To compare and contrast them, we first briefly introduce the structure of SGNS and then introduce the proposed TSGNS.

Figure \ref{fig:TSGNS} (a) shows the diagram of SGNS for time period $t$ \cite{Rong:2016}. The input is the one-hot representation of individual words in a vocabulary; for example, the one-hot of word $w$ is a vector $\{0,0,1,\cdots,0\}$ where $1$ indicates the position corresponding to word $w$ and $0$ for all other words. The output $y_{C,j}$ is the probability that word $j$ appears at the $C^{th}$ position in the context of the input word. Given the context range $L$, one has $C=2L$. For example, if the context range $L=2$, the two words before and after the input word are in the $1^{st}$, $2^{nd}$, $3^{rd}$ and $4^{th}$ position respectively in the context.

After training SGNS using the corpus segment in time period $t$, $\mathbf{W}^{(t)}_{V\times N}$ is the matrix where the embedding of word $w$ (i.e., $\vec{w}^{(t)}$) is the row corresponding to $w$ in $\mathbf{W}^{(t)}_{V\times N}$, that is, the weights of links from node $w$ in the input layer to all nodes in the hidden layer;

Note that $\vec{w}^{(t)}$ can be viewed as a vector in the $N$-dimensional space defined by the hidden layer. In turn, the $N$-dimensional space can be viewed as a subspace in the $(C\times V)$-dimensional space defined by the output layer. Specifically, the $i^{th}$ node in the the hidden layer, denoted as $h_i$, is a vector in the $(C\times V)$-dimensional space. The vector consists of the values in $i^{th}$ row in matrix $\mathbf{X}^{(t)}_{N\times V,1}$, $\mathbf{X}^{(t)}_{N\times V,2}$, $\cdots$, $\mathbf{X}^{(t)}_{N\times V, C}$, that is, the weights of links from the $i^{th}$ node in the hidden layer (i.e., $h_i$) to all nodes in the output layer.

For time period $t+1$, another model is trained in the similar way as for time period $t$. Typically, the corpus segment in time period $t$ differs from that in time period $t+1$; and thus matrix $\mathbf{X}^{(t+1)}_{N\times V,1}$, $\mathbf{X}^{(t+1)}_{N\times V,2}$, $\cdots$, $\mathbf{X}^{(t+1)}_{N\times V, C}$ will be different from matrix $\mathbf{X}^{(t)}_{N\times V,1}$, $\mathbf{X}^{(t)}_{N\times V,2}$, $\cdots$, $\mathbf{X}^{(t)}_{N\times V, C}$. As a result, the embedding of $w$ in time period $t$ and the embedding of $w$ in time period $t+1$ cannot be comparable properly.

The scheme of TSGNS is to attach each word $w$ in the input layer with a tag to indicate the time period associated (i.e., $w^{(t)}$ and $w^{(t+1)}$ in time period $t$ and $t+1$ respectively) and they are then embedded in a common vector space.

The structure of TSGNS is illustrated in Figure \ref{fig:TSGNS} (b) which incorporates both time period $t$ and $t+1$. Suppose the one-hot representation of $w$ is $\{0,0,1,\dots,0\}$, the one-hot representation of $w^{(t)}$ is $\{0,0,1,\dots,0\}\oplus$ $\{0,0,0,\dots,0\}$ and that of $w^{(t+1)}$ is $\{0,0,0,\dots,0\}\oplus$ $\{0,0,1,\dots,0\}$ where $\oplus$ is the vector concatenation operator.

After training TSGNS with the corpus segments in time period $t$ and $t+1$, the word embedding of $w^{(t)}$ ($\vec{w}^{(t)}$) is the row in matrix $\mathbf{W}^{(t)}_{V\times N}$ corresponding to $w^{(t)}$, i.e., the weights of links from the node $w^{(t)}$ in the input layer to all nodes in the hidden layer; and the word embedding of $w^{(t+1)}$ ($\vec{w}^{(t+1)}$) is the row of matrix $\mathbf{W}^{(t+1)}_{V\times N}$ corresponding to $w^{(t+1)}$, i.e., the weights of links from the node $w^{(t+1)}$ in the input layer to all nodes in the hidden layer.

Different from SGNS, the word embeddings $\vec{w}^{(t)}$ and $\vec{w}^{(t+1)}$ both are projected in the same $N$-dimensional subspace defined by the hidden layer in the $(C\times V)$-dimensional space defined by the output layer.

\subsection{Optimize Tagged-SGNS}
As shown in Figure \ref{fig:TSGNS} (b), the input of TSGNS is $V2 = [V, V]$. Similar to SGNS, training word embedding in TSGNS across time are optimized by maximizing the average log probability.

\begin{equation}\label{eq:opt}
\displaystyle \mathop{\arg\max}\Big\{\frac{1}{|V2|}\sum_{i=1}^{|V2|}\sum_{-L\leqslant j\leqslant L,j\neq 0,}\log p(w_{i+j}|w_i)\Big\}
\end{equation}
where $w_i$ is the input word, $L$ is the context distance, $w_{i+j}$ is the word in $j^{th}$ position in the context of $w_i$, $p(*|*)$ is a softmax function:
\begin{equation}\label{eq9}
\displaystyle p(w_{i+j}|w_i)=\frac{\exp(\vec w_{i+j}^\top \vec w_i)}{\Sigma_{x=1}^{|V2|}\exp(\vec w_{x}^\top \vec w_i)}
\end{equation}
where $\vec w_{*}^\top \vec w_i$ is the value of node $w_{*}$ in the output layer, and $\vec w_{*}$ is the vector which consists of the weights of links from all nodes in the hidden layer to node $w_*$ in the output layer.

\subsection{Embedding Similarity Measure}\label{sec:sim}
Suppose words have been embedded in the same vector space across time using TSGNS. Given word $w$, the similarity of word embeddings in two time periods ($\vec{w}^{(t)}$ and $\vec{w}^{(t+1)}$) can be measured using cosine similarity, defined as follows:
\begin{equation}\label{eq8}
sim(w|D^{(t)}, D^{(t+1)})=cossim({\vec{w}^{(t)}}, {\vec{w}^{(t+1)}})
\end{equation}
where $D^{(t)}$ and $D^{(t+1)}$ are the corpus segments in time period $t$ and $t+1$ respectively. The value of $sim(w|D^{(t)}, D^{(t+1)})$ is in [0,1]. The smaller value means they are less similar.

\subsection{Smooth Alignment Property of TSGNS}
This section discusses the smooth alignment property of TSGNS.

\begin{property}
Given the corpus segments in time period $t$ and $t+1$, the word embedding generated using TSGNS is of smooth alignment.
\end{property}

\begin{figure}[t]
	\centering
	\includegraphics[width=0.5\textwidth]{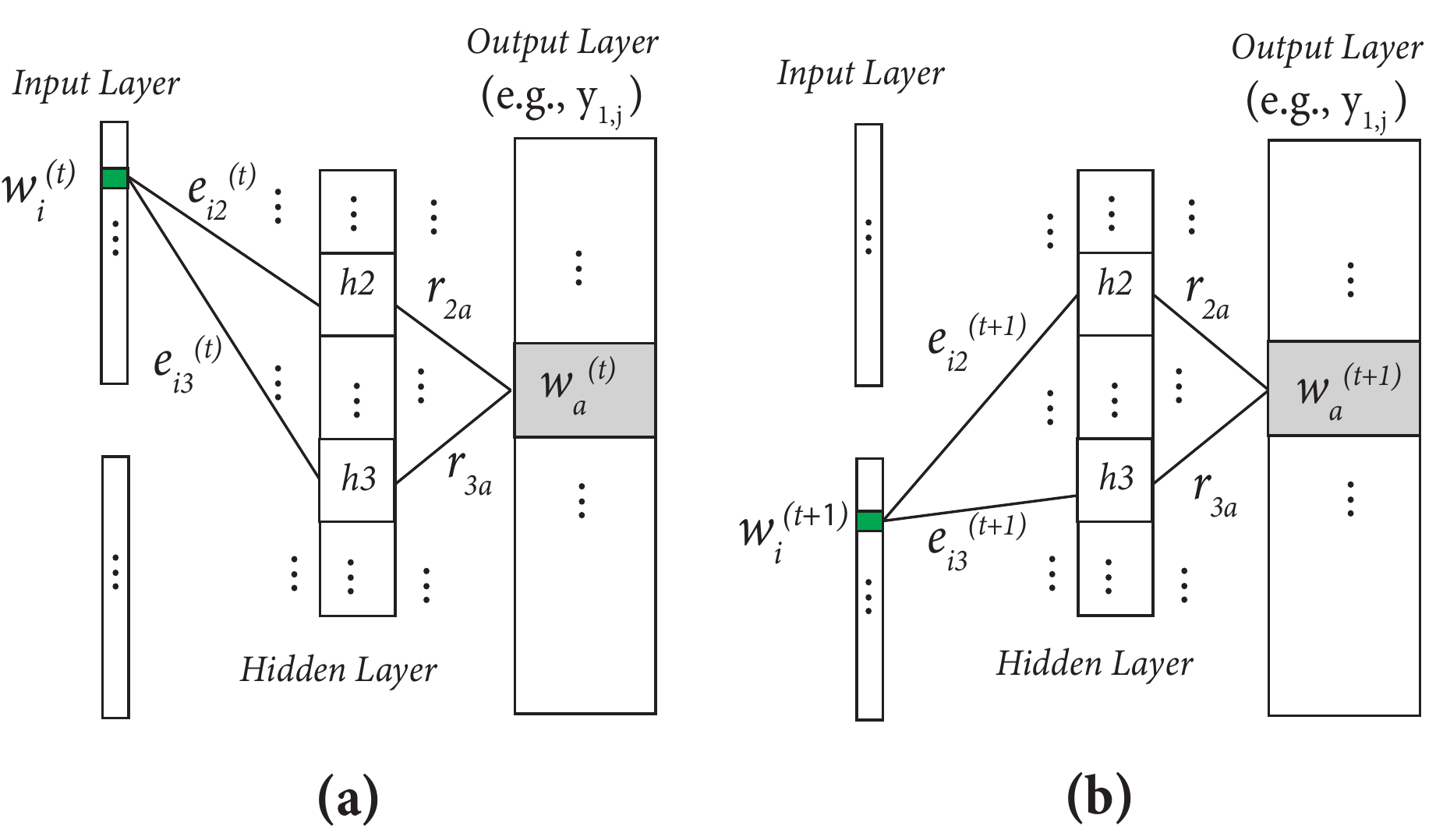}
	\caption{$w_a^{(t)}$ is the probability that word $w_a$ is in the context of word $w_i$ in time periods $t$, so is $w_a^{(t+1)}$ in time periods $t+1$.}
	\label{fig:explanation}
\end{figure}

We now show the smooth alignment property of TSGNS with help of Figure \ref{fig:explanation}. For a word $w_i$, it is denoted as $w_i^{(t)}$ in time period $t$ and as $w_i^{(t+1)}$ in time period $t+1$ respectively. Without loss of generality, Figure \ref{fig:explanation} only shows $y_{1,j}$ in output layer of TSGNS for presentation simplicity. 

As shown in Figure \ref{fig:explanation} (a)(b), $w_a^{(t)}$ is the probability that word $w_a$ is in the context of word $w_i$ in time periods $t$, so is $w_a^{(t+1)}$ in time periods $t+1$. We would like to reveal that (1) if $w_a^{t}$ is more similar to $w_a^{(t+1)}$, the word embedding of $w_i$ in time period $t$ (i.e., $\vec{w}_i^{(t)}$) and the word embedding of $w_i$ in time period $t+1$ (i.e., $\vec{w}_i^{(t+1)}$) tend to be more similar; (2) if $w_a^{t}$ is more dissimilar to $w_a^{(t+1)}$, $\vec{w}_i^{(t)}$) and $\vec{w}_i^{(t+1)}$) tend to be less similar.

As discussed in Section \ref{sec:tsgns}, a $(C\times V)$-dimensional space, denoted as $\mathbb{F}$, is defined by the output layer where each dimension corresponds to a node, i.e., a context word in the output layer. The word embedding $\vec{w}_i^{(t)}$ is $[e_{i1}^{(t)},\cdots,e_{ij}^{(t)},\cdots]$ where $e_{ij}^{(t)}$ is the weight of link from node $w_i^{(t)}$ in the input layer to node $h_j$ in the hidden layer. 

When $w_i^{(t)}$ is the input (i.e., in the input layer, only the node corresponding to $w_i^{(t)}$ is 1, all other nodes are 0), the value of node $w_a^{(t)}$ in the output layer is $\sum_{h_j\in H}(r_{ja} e_{ij}^{(t)})$ where $H$ is the set of all nodes in the hidden layer. When $w_i^{(t+1)}$ is the input, the value of node $w_a^{(t+1)}$ in the output layer is $\sum_{h_j\in H}(r_{ja} e_{ij}^{(t+1)})$. 

If the value of node $w_a^{(t)}$ is more similar to that of $w_a^{(t+1)}$,
\begin{equation}\label{eq:wa}
\lim_{w_a^{(t)} \to w_a^{(t+1)}} \sum_{h_j\in H}(r_{ja} e_{ij}^{(t)})=\sum_{h_j\in H}(r_{ja} e_{ij}^{(t+1)})
\end{equation}
If there are more context words like $w_a$, there are more equations like Eq. (\ref{eq:wa}), each for one such context word. In this situation, training TSGNS will enforce the distance between values of $\vec{w}_i^{(t)}$ and $\vec{w}_i^{(t+1)}$ tends to decrease such that $\vec{w}_i^{(t)}$ and $\vec{w}_i^{(t+1)}$ tends to be more similar according to Eq. (\ref{eq8}).

\subsection{Discussion}\label{sec:discussion}
\subsubsection{Multiple Time Periods} 
Word embedding using TSGNS across two time periods has been discussed. It is easy to extend TSGNS for word embedding across more than two time periods while the property of smooth alignment holds. To this end, the output layer and hidden layer shown in Figure \ref{fig:TSGNS} (b) remain the same, but the input layer is extended from $2\times V$ to $T\times V$ where $T$ is the number of time periods. The smooth alignment can be verified in the similar way as in the case of two time periods, i.e., we can still guarantee smooth alignment in $T$ time periods of embedding.

\subsubsection{Tagged-SVD}
As discussed in Section \ref{sec:related1}, there are two main low-dimensional embedding methods, i.e., SGNS and SVD. In addition to SGNS, the time-tagged word embedding scheme can also be applied to SVD (called \textit{Tagged-SVD} (TSVD) following the name convention of TSGNS). In this situation, the PPMI matrix is extended.
\begin{equation}
PPMI(<t,t+1>,L)=[PPMI(t,L), PPMI(t+1,L)]
\end{equation}
where $L$ is the context range. Each word has two rows in $PPMI(<t,t+1>,L)$ to represent its co-occurrence statistics in $D^{(t)}$ and $D^{(t+1)}$ respectively. The word embeddings can be obtained by following Eq. (\ref{eq:svd}). Similar to TSGNS, words across time are embedded in a common vector space defined by $\mathbf{I}_d^{(t,t+1)}$. Also, it is straightforward to extend PPMI across any number of time periods to be $PPMI(<\cdots>,L)$ and then apply SVD. In the similar way as for TSGNS, we can also prove that the embedding using TSVD is of smooth alignment.

However, TSVD optimization requires processing $PPMI(<...>,L)$ which can be too large to fit in memory. So, this study focuses on TSGNS only.

\section{Experimental Results and Analysis}\label{sec:test}
\subsection{Experimental Settings}
Google Books is a web service that searches the full text of books and magazines that Google has scanned, converted to text using optical character recognition (OCR), and stored in its digital database. This paper uses an 105GB dataset from Google Books N-gram (version: English 20120701, type: 5-gram)~\footnote{http://storage.googleapis.com/books/ngrams/books/datasetsv2.html.}. The corpus describes the phrases used from 1900 to 2000. In total there are 250 millions unique 5-grams and, due to duplication, the total number of all 5-grams (or lines) is much more.

\begin{figure}[t]
	\centering
	\includegraphics[width=1\linewidth]{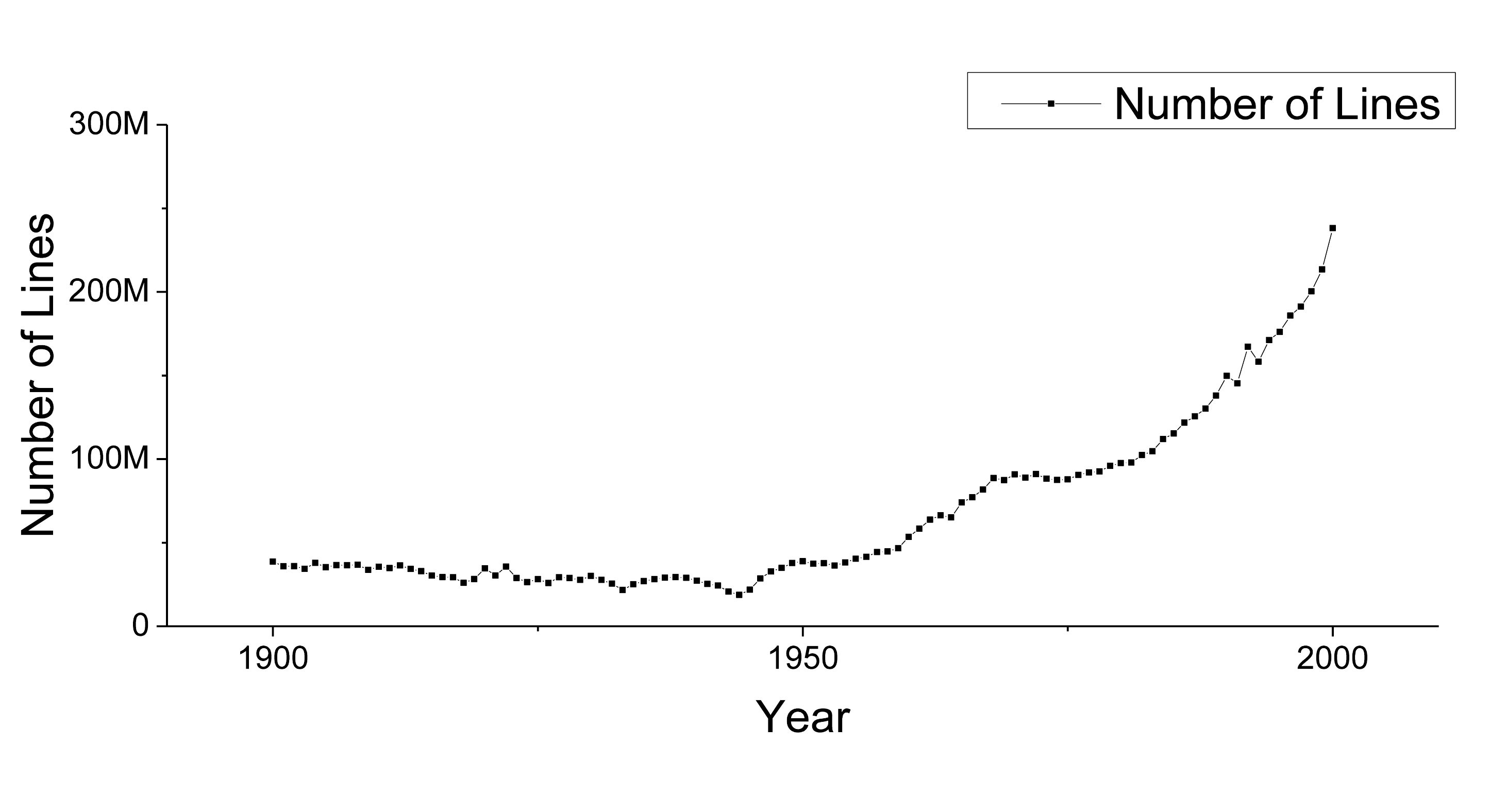}
	\caption{The number of Google Books 5-grams by year.}
	\label{fig:datayear}
\end{figure}

We compare the proposed TSGNS against PPMI, SVD, SGNS \cite{hamilton2016diachronic}, and DW2V \cite{Yao:2018}. The hidden layer of TSGNS contains 300 nodes. The time from 1900 to 2000 is uniformly split by year. Accordingly, all 5-grams in the dataset are split into corpus segments, each for one year. After the split, the dataset in each year is still dense. Figure \ref{fig:datayear} shows the number of lines in each year. It shows that the number of lines increases dramatically after 1950s. The cosine similarity (Eq. (\ref{eq8})) is used to measure the distances between word embeddings.

\begin{figure}[h]
	\centering
	\includegraphics[width=1.08\linewidth]{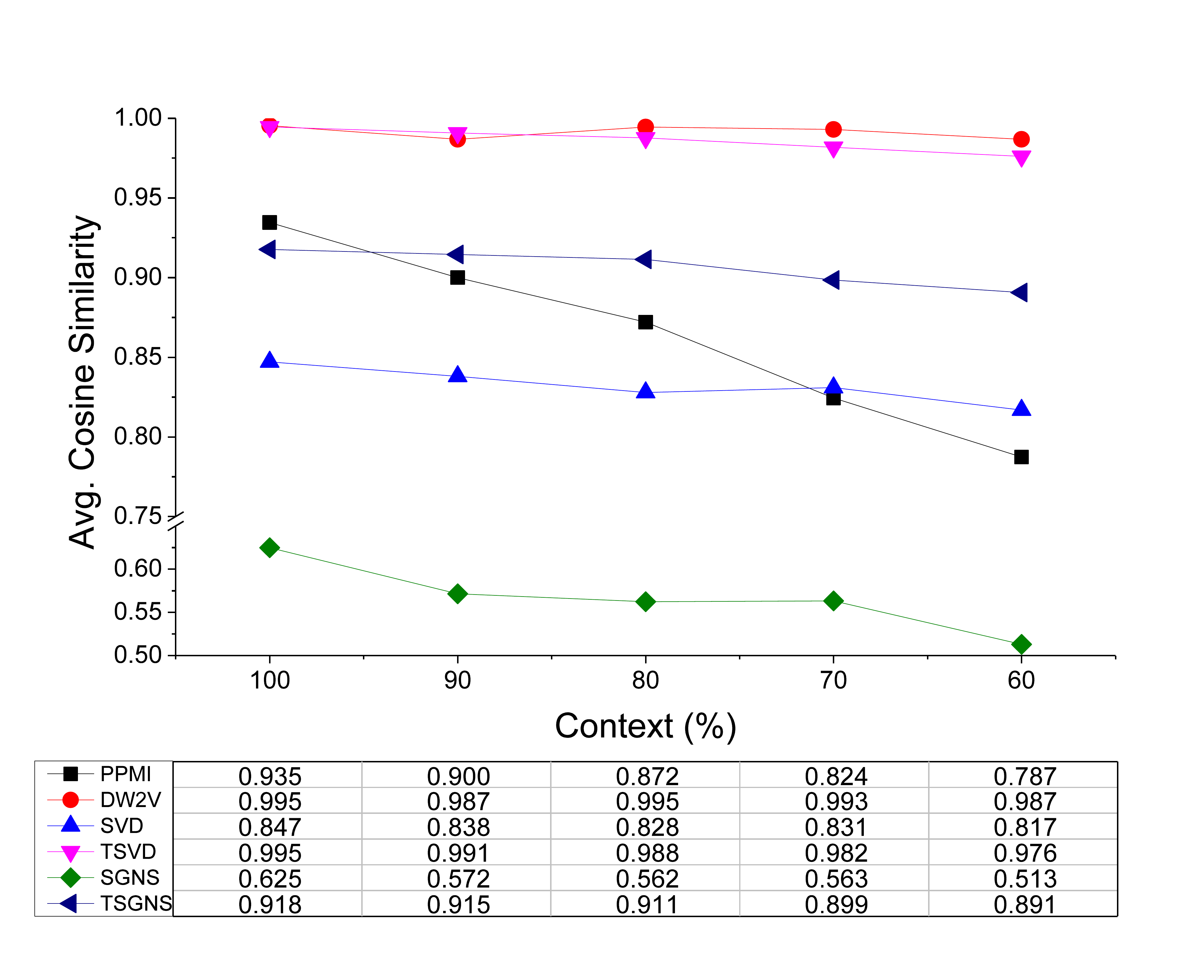}
	\caption{When context overlap percentage decreases (i.e., co-occurrence statistics are less similar), average cosine similarity of word embeddings using TSGNS and TSVD decreases monotonically.}
	\label{fig:trend}
\end{figure}

\begin{table*}
	\setlength{\abovecaptionskip}{0pt}
	\setlength{\belowcaptionskip}{0pt}
	\caption{Cosine similarity of word embeddings on selected words.}
	\centering
    \begin{tabular}{c|D{\%}{\%}{4.1}|d{3.4}|d{3.4}|d{3.4}|d{3.4}|d{3.4}|d{3.4}|d{3.4}|d{3.4}|d{3.4}|d{3.4}}
		\hline
        \textbf{} & \multicolumn{1}{c|}{\textbf{Context}} & \multicolumn{1}{c|}{\textbf{mirage}} & \multicolumn{1}{c|}{\textbf{\small{localized}}} & \multicolumn{1}{c|}{\textbf{lord}} & \multicolumn{1}{c|}{\textbf{\small{recollections}}} & \multicolumn{1}{c|}{\textbf{tired}} & \multicolumn{1}{c|}{\textbf{thunder}} & \multicolumn{1}{c|}{\textbf{\small{reporter}}} & \multicolumn{1}{c|}{\textbf{bridge}} & \multicolumn{1}{c|}{\textbf{web}} & \multicolumn{1}{c}{\textbf{reject}} \\
		\hline
		\multirow{4}{*}{PPMI} & 100 \% & 0.937 & 0.909 & 0.972 & 0.978 & 0.951 & 0.79 & 0.926 & 0.888 & 0.996 & 0.999\\ \cline{2-12}
		& 90 \% & 0.921 & 0.883 & 0.93 & 0.925 & 0.91 & 0.777 & 0.871 & 0.859 & 0.972 & 0.952\\ \cline{2-12}
		& 80 \% & 0.882 & 0.864 & 0.887 & 0.899 & 0.854 & 0.768 & 0.838 & 0.835 & 0.944 & 0.95\\ \cline{2-12}
		& 70 \% & 0.852 & 0.846 & 0.847 & 0.859 & 0.783 & 0.758 & 0.791 & 0.791 & 0.904 & 0.813\\ \cline{2-12}
        & 60 \% & 0.822 & 0.828 & 0.796 & 0.819 & 0.784 & 0.734 & 0.774 & 0.739 & 0.879 & 0.699\\ \hline
		\multirow{4}{*}{DW2V} & 100 \% & 0.993 & 0.992 & 0.998 & 0.996 & 0.997 & 0.992 & 0.999 & 0.988 & 0.999 & 1.0\\ \cline{2-12}
		& 90 \% & 0.995 & 0.999 & 1.0 & 0.986 & 0.999 & 0.997 & 0.964 & 0.999 & 0.95 & 0.979\\ \cline{2-12}
		& 80 \% & 0.992 & 0.986 & 0.995 & 0.992 & 1.0 & 0.988 & 0.995 & 1.0 & 0.998 & 0.999\\ \cline{2-12}
		& 70 \% & 0.986 & 0.973 & 0.999 & 0.991 & 0.998 & 0.999 & 0.997 & 0.99 & 1.0 & 0.997\\ \cline{2-12}
        & 60 \% & 0.99 & 0.968 & 0.999 & 0.995 & 1.0 & 0.926 & 0.996 & 0.998 & 1.0 & 0.996\\ \hline
		\multirow{4}{*}{SVD} & 100 \% & 0.868 & 0.867 & 0.856 & 0.858 & 0.833 & 0.856 & 0.809 & 0.866 & 0.802 & 0.856\\ \cline{2-12}
		& 90 \% & 0.861 & 0.877 & 0.857 & 0.844 & 0.819 & 0.848 & 0.775 & 0.85 & 0.788 & 0.861\\ \cline{2-12}
		& 80 \% & 0.849 & 0.874 & 0.837 & 0.834 & 0.833 & 0.848 & 0.748 & 0.861 & 0.78 & 0.815\\ \cline{2-12}
		& 70 \% & 0.868 & 0.861 & 0.851 & 0.826 & 0.787 & 0.848 & 0.713 & 0.885 & 0.817 & 0.854\\ \cline{2-12}
        & 60 \% & 0.844 & 0.862 & 0.833 & 0.844 & 0.798 & 0.806 & 0.739 & 0.836 & 0.825 & 0.782\\ \hline
		\multirow{4}{*}{TSVD} & 100 \% & 0.993 & 0.993 & 1.0 & 0.998 & 0.997 & 0.989 & 0.983 & 0.994 & 1.0 & 0.999\\ \cline{2-12}
		& 90 \% & 0.99 & 0.992 & 0.999 & 0.997 & 0.988 & 0.988 & 0.973 & 0.991 & 0.999 & 0.991\\ \cline{2-12}
		& 80 \% & 0.988 & 0.99 & 0.997 & 0.992 & 0.98 & 0.988 & 0.964 & 0.988 & 0.998 & 0.992\\ \cline{2-12}
		& 70 \% & 0.984 & 0.988 & 0.995 & 0.986 & 0.967 & 0.986 & 0.951 & 0.983 & 0.998 & 0.98\\ \cline{2-12}
        & 60 \% & 0.981 & 0.985 & 0.993 & 0.98 & 0.958 & 0.986 & 0.932 & 0.978 & 0.997 & 0.971\\ \hline
		\multirow{4}{*}{SGNS} & 100 \% & 0.626 & 0.706 & 0.707 & 0.542 & 0.489 & 0.667 & 0.614 & 0.672 & 0.723 & 0.502\\ \cline{2-12}
		& 90 \% & 0.648 & 0.67 & 0.654 & 0.486 & 0.451 & 0.619 & 0.435 & 0.57 & 0.681 & 0.501\\ \cline{2-12}
		& 80 \% & 0.579 & 0.66 & 0.663 & 0.52 & 0.49 & 0.575 & 0.477 & 0.606 & 0.702 & 0.352\\ \cline{2-12}
		& 70 \% & 0.587 & 0.667 & 0.63 & 0.536 & 0.484 & 0.526 & 0.456 & 0.666 & 0.712 & 0.369\\ \cline{2-12}	
        & 60 \% & 0.467 & 0.605 & 0.631 & 0.513 & 0.507 & 0.503 & 0.41 & 0.591 & 0.653 & 0.25\\ \hline
		\multirow{4}{*}{TSGNS} & 100 \% & 0.942 & 0.924 & 0.883 & 0.931 & 0.897 & 0.914 & 0.947 & 0.859 & 0.918 & 0.962\\ \cline{2-12}
		& 90 \% & 0.934 & 0.924 & 0.872 & 0.934 & 0.913 & 0.908 & 0.945 & 0.859 & 0.914 & 0.942\\ \cline{2-12}
		& 80 \% & 0.922 & 0.936 & 0.891 & 0.926 & 0.897 & 0.912 & 0.928 & 0.863 & 0.903 & 0.936\\ \cline{2-12}
		& 70 \% & 0.914 & 0.907 & 0.859 & 0.905 & 0.895 & 0.909 & 0.919 & 0.851 & 0.899 & 0.927\\ \cline{2-12}
        & 60 \% & 0.903 & 0.91 & 0.859 & 0.897 & 0.892 & 0.898 & 0.909 & 0.851 & 0.882 & 0.905\\ \hline	
	\end{tabular}\label{tbl:align}
\end{table*}

\subsection{Alignment Smoothness}
To verify the smooth alignment of TSGNS, we test a number of randomly selected words in time period $t$ as input word where $t$ is any year from 1900 to 1999. For each input word $w$ in the next time period $t+1$, we first remove all lines (i.e., 5-grams) related to $w$ in the corpus segment, i.e., the lines where word $w$ is in the middle of the 5-grams; then we copy all lines related to $w$ in the corpus segment in time period $t$ to the corpus segment in time period $t+1$. By manipulating in such way, $w$ has the exactly same co-occurrence statistics (denoted as $100\%$) in time period $t$ and $t+1$.

Moreover, for each input word $w$, we also select $\alpha$ percentage of all lines related to $w$ in time period $t+1$ and replace its context words using the randomly select words in the vocabulary. The setting of $\alpha$ is $0\%$, $10\%$, $20\%$, $30\%$, and $40\%$ such that the co-occurrence statistics of $w$ in time period $t$ and $t+1$ are overlapped by $100\%$, $90\%$, $80\%$, $70\%$, and $60\%$ respectively.

At different settings of $\alpha$, TSGNS is trained using the corpus segments in $t$ and $t+1$. For each input word, the word embeddings in the two time periods are generated and their cosine similarity is computed. Ten different words are selected. For each word in time period $t=1989$ and $t+1=1990$, the cosine similarities at different settings of $\alpha$ are computed using PPMI, SGNS, SVD, DW2V, TSGNS, and TSVD respectively. The test results are reported in Table \ref{tbl:align}. To make it clear in comparison, the average cosine similarity on the ten words for each method is calculated and presented in Figure \ref{fig:trend}. 

We observe that the average cosine similarity consistently decreases using TSGNS and TSVD when $\alpha$ changes from $100\%$ to $60\%$. In contrast, the average cosine similarity does not consistently decrease using SGNS, SVD, and DW2V. It verifies the smooth alignment property of TSGNS and TSVD. 

It is worthy to point out that the absolute value of cosine similarity does not make much sense. In contrast, it is essential that the cosine similarity can gauge to which extent that the co-occurrence statistics differ.

Figure \ref{fig:trend} verifies that PPMI is smooth aligned by nature. As discussed, however, PPMI does not enjoy the advantages of low-dimensional embeddings such as higher efficiency and better generalization. 

\subsection{Synchronic Accuracy}
Synchronic linguistics is the study of the linguistic elements and usage of a language at a particular moment.

\subsubsection{Semantic Similarity}
The words known with similar semantic should have the similar word embeddings. In this test, the words with known similar semantics are from Bruni et al.'s MEN similarity task of matching human judgments of word similarities \cite{bruni2012distributional}. The total number of such word pairs is 3000. 

We tested semantic similarity in both short term and long term. For the short term, each time period covers a year ($t=1989$, $t+1=1990$); and for the long term, each time period covers ten years ($t=1980$-$1989$, $t+1=1990$-$1999$). For each word, the word embeddings based on the corpus segment covering the two time periods are generated using SGNS and SVD; the word embeddings are generated using TSGNS and TSVD where $t$ is the first time period and $t+1$ is the next time period. For the 3000 words, we test the Spearman's correlation between the word embedding similarities and human judgments \cite{hamilton2016diachronic}. The results are presented in Table \ref{tbl:semantic}. TSGNS and SGNS have comparable performance; TSVD and SVD have comparable performance. The results verify that TSGNS and TSVD can properly measure the semantic similarity at a particular time period, although they are designed for word embedding across time.


\begin{table}
	\setlength{\abovecaptionskip}{0pt}
	\setlength{\belowcaptionskip}{0pt}
	\caption{Ability to capture semantic similarity.}
	\centering
	\begin{tabular}{c|c|c||c|c}
		\hline
		\textbf{} & \textbf{Methods} & \textbf{Correlation} & \textbf{Methods} & \textbf{Correlation}  \\
		\hline
		\textbf{Short} 	& SVD    &  0.590 & SGNS   & 0.284 \\ \cline{2-5}
		\textbf{Term}   & TSVD  &  0.602 & TSGNS & 0.294\\ \hline \hline
		\textbf{Long}  	& SVD    &  0.633 & SGNS   & 0.288 \\ \cline{2-5}
		\textbf{Term}  	& TSVD  &  0.590 & TSGNS & 0.349\\ \hline
	\end{tabular}\label{tbl:semantic}
\end{table}


\begin{table}
	\setlength{\abovecaptionskip}{0pt}
	\setlength{\belowcaptionskip}{0pt}
	\caption{Average correlation between embedding norm and normalized frequency of selected words.}
	\centering
	\begin{tabular}{c|c||c|c}
		\hline
		\textbf{Methods} & \textbf{Correlation} & \textbf{Methods} & \textbf{Correlation}  \\
		\hline
		PPMI   &  0.609 & SVD  & 0.248 \\  \hline
		SGNS    &  0.197 & TSGNS   & 0.577 \\ \hline
	\end{tabular}\label{tbl:vecnorm}
\end{table}

\begin{table}[h]
    \setlength{\abovecaptionskip}{0pt}
	\setlength{\belowcaptionskip}{0pt}
	\caption{Ability to capture the attested shifts.}
	\centering
	\begin{tabular}{c|c|c||c|c}
		\hline
						& \textbf{Methods} & \textbf{Correlation} & \textbf{Methods} & \textbf{Correlation}  \\
		\hline
		\textbf{Short} 		&PPMI   &  0.423 & DW2V  &  N/A  \\  \cline{2-5}
		\textbf{Term}			& SVD    &  0.146 & SGNS  & 0.075 \\ \cline{2-5}
											& TSVD   &   N/A  & TSGNS & 0.347 \\ \hline \hline
		\textbf{Long}		&PPMI   &  0.305 & DW2V  &  0.141  \\  \cline{2-5}
		\textbf{Term}			&SVD    &  0.395 & SGNS  & 0.069 \\ \cline{2-5}
											&TSVD   &   0.337  & TSGNS & 0.485 \\ \hline
	\end{tabular}\label{tbl:known}
\end{table}

\begin{table}
\setlength{\abovecaptionskip}{0pt}
\setlength{\belowcaptionskip}{0pt}
\caption{Top-10 words with the highest semantic displacement values from 1900 to 2000.}
\centering
\begin{tabular}{c|p{6.5cm}}
\hline
\textbf{Methods} & \textbf{Discovered words} \\
\hline
PPMI & greed \textbf{landowner} \underline{dating} panacea investigator \textbf{mobile} donation \textbf{flicker} bonfire badge \\
\hline
SVD & bonfire \underline{fixture} horde spin \textbf{textbook} passer facility \textbf{broadway} \textbf{ flicker} bulwark \\
\hline
SGNS & \textbf{gay} \textbf{thrust} \textbf{van} pearl fault \textbf{smoking} tear approach sink magnet \\
\hline
TSGNS & approach \textbf{display} \textbf{album} \textbf{publishing} \underline{signal} \textbf{gay} \underline{economy} major demonstration \textbf{van} \\
\hline
\end{tabular}\label{tbl:top10}
\end{table}

\begin{table*}
	\setlength{\abovecaptionskip}{0pt}
	\setlength{\belowcaptionskip}{0pt}
	\caption{Context shift of words from 1980 to 2000.}
	\centering
	\begin{tabular}{c|c|c|c}
		\hline
		\textbf{Methods} & \textbf{computer} & \textbf{earthquake} & \textbf{microsoft} \\
		\hline
		PPMI &  microcomputer (80) PC (85) internet (94) & prieta (90) awaji (96) & visicorp (83) wordperfect (86) compuserve (98) \\
		\hline
		SVD &  digital (80) software (85) internet (95) & alcatraz (90) prieta (97)& zilog (83) wordperfect (87) macromedia (97) \\
		\hline
		SGNS & model (80) application (93) modem (99) & nevada (84) sudan (91) & dell (87) unix (92) \\
		\hline
		TSGNS & modem (82) programming (88) desktop (93) & hiroshima (91) prieta (95) & xerox (88) unix (91) netscape (98) \\
		\hline
	\end{tabular}\label{tbl:context}
\end{table*}

\begin{figure*}[h!]
	\centering
        \includegraphics[width=0.3\textwidth]{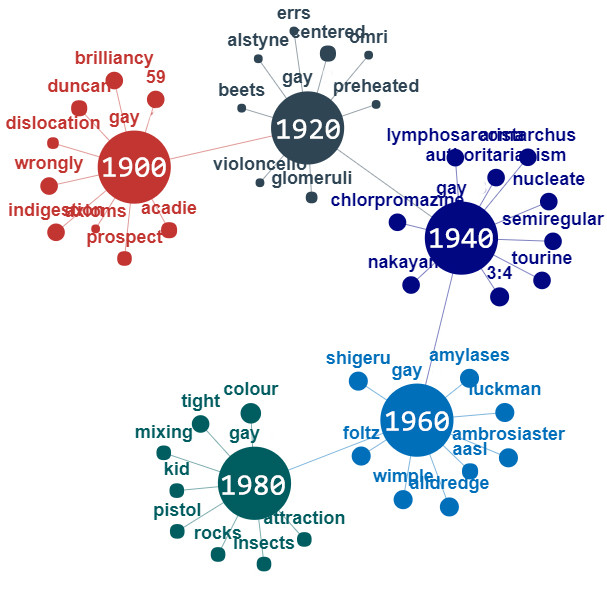}
        \label{fig:visual1}
        \includegraphics[width=0.3\textwidth]{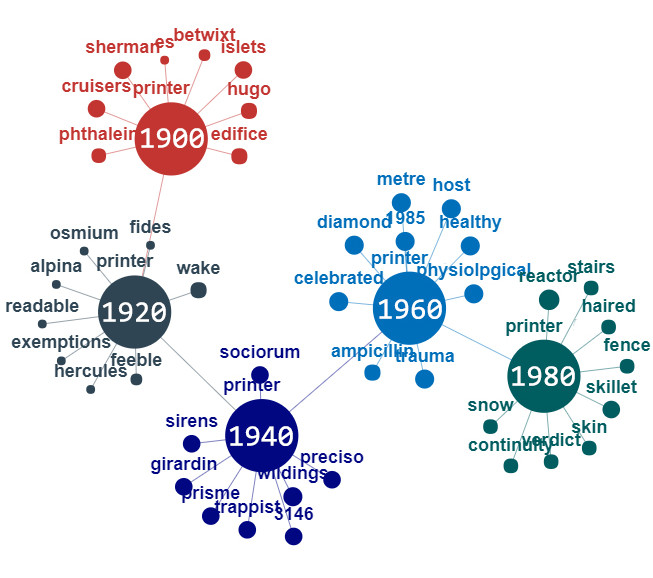}
        \label{fig:visual2}
        \includegraphics[width=0.3\textwidth]{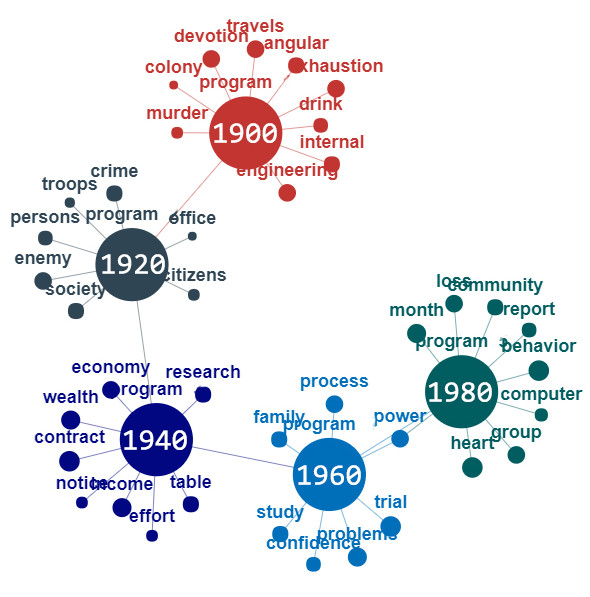}
        \label{fig:visual3}
\caption{Visualization of semantic change of words (a) \textit{gay}, (b) \textit{printer} and (c) \textit{program}.}\label{fig:shifts}
\end{figure*}

\subsubsection{Vector Norm vs. Frequency}
It has been observed that word embeddings computed by factorizing PPMI matrices have norms that grow with word frequency \cite{Yao:2018}. These word vector norms can be viewed as a time series for detecting the trend concepts behind words with more robustness than word frequency. Here, we test the Spearman's correlation between the embedding norm of 10 randomly selected words and the normalized frequency of corresponding words along time from 1900 to 1999. Given a word, the normalized frequency is the number of times it appears in a corpus segment normalized by the total number of words in the same corpus segment. Table~\ref{tbl:vecnorm} illustrates the average correlation between embedding norm and normalized frequency of selected words. We can observe that the TSGNS has word embedding norm more consistent with the normalized frequency than other low-dimensional embedding methods, and has a similar performance as PPMI.

\subsection{Diachronic Validity}
Diachronic linguistics studies the changes in language over time.

\subsubsection{Semantic Change}\label{sec:semchange}
When the semantic change of words happens, the capability of methods to capture the shifts is tested.

\textit{Verify Against Known Shifts.} We have tested the capacity of different methods to capture known historical shifts in meaning. That is, the cosine similarity between word embeddings generated can correctly capture whether pairs of words moved closer or further apart in semantic space, or the pairwise similarity series have the correct sign on their Spearman correlations. We evaluated the methods on a dataset from Oxford Dictionaries\footnote{https://en.oxforddictionaries.com/explore/archaic-words/}. The dataset contains 412 words which are human-recognised with semantic shift over time. On this dataset, we test the proposed methods against all baseline methods (PPMI, SVD, SGNS and DW2V). The word embeddings across 100 years are generated using TSGNS in short term (each time period covering one year, i.e., 1900, 1901, $\cdots$, 1999) and long term (each time period covering ten years, i.e., 1900-1909, 1910-1919, $\cdots$, 1990-1999). The input layer is extended from 2$\times V$ to $T\times V$ where $T=100$ in short term or $T=10$ in long term. We cannot get result using TSVD and DW2V in short term setting due to huge requirement of memory. 

The Spearman correlations of methods are shown in Table \ref{tbl:known}. Clearly, TSGNS beats SGNS and SVD, and has similar performance as PPMI. It is worth pointing out that PPMI does not enjoy the advantages of low-dimensional embeddings such as higher efficiency and better generalization. In specific, the low-dimensional embeddings are influenced more or less by the context change of all words over time; on the other hand, the high-dimensional embeddings using PPMI are isolated from word to word, i.e., the embedding of a word is not influenced by the context change of other words over time. It helps explain why PPMI performs even worse in long term since it is expected that context change in long term is greater than in short term.




\textit{Discovering Shifts from Data.} We have tested whether the methods discover reasonable shifts by examining the top-10 words that changed the most from 1900 to 2000. The embeddings are generated in the same way as in the test \textit{Verify Against Known Shifts} (long term). Table \ref{tbl:top10} shows the top 10 words discovered by each method. These shifts have been judged by authors as being either clearly genuine (bold), borderline (underline) or clearly corpus artifacts. Interestingly, TSGNS demonstrates good performance since it can identify words known changed in the past years such as \textit{van}, \textit{album}, \textit{display} while other methods cannot.


\subsubsection{Context Change}
Another application of word embedding alignment across time is to help identify the conceptually equivalent items or people over time. This section provides examples in the field of \textit{technology} (i.e., computer), \textit{natural phenomena} (i.e., earthquake), and \textit{well-known business} (i.e., Microsoft). In the test, we create a query consisting of a word-year pair that is particularly the representative of that word in that year, and looking for other word-year pairs in its vicinity in different years. The word embeddings across 20 years (1980-2000) are generated using TSGNS where the input layer is $20\times V$. For the same reason as discussed in Section \ref{sec:semchange}, we cannot get results using TSVD and DW2V. The test outputs are presented in Table \ref{tbl:context}. Comparing with PPMI, SVD and SGNS, we believe that the outputs of TSGNS make more sense, e.g., \textit{desktop (93)} and \textit{programming (88)} are more relevant to computer than \textit{model (80)} and \textit{application (99)}.

\textit{Shift Visualization.} Visualizing trajectories of word over time is intuitive to reveal the semantic shift of words. Figure \ref{fig:shifts} shows the trajectories of three words in 1900, 1920, 1940, 1960 and 1980 respectively based on the outputs of TSGNS. For each word in a year, the closest words in terms of cosine similarity are attached. In Figure \ref{fig:shifts} (a), the semantic of word \emph{gay} shifted from ``brilliancy'' (during year 1900-1919) to ``pistol'' or ``attraction'' (during year 1980-2000); (b) the semantic of word \emph{printer} shifted from ``edifice'', ``fides'' to referring to a household equipment like ``skillet"; (c) the semantic of word \emph{program} was related to the military as ``colony'' (during year 1900-1919) or ``troops'' (during year 1920-1939), and then its meaning changed to economic activities as ``economy'', ``contract'' or ``wealth'' (during year 1940-1959). With the rise of computer science its meaning got close to ``computer'' or ``report'' (during year 1980-2000). 

\section{Conclusion}\label{sec:conclusion}
The proposed TSGNS has addressed the alignment problem of vector spaces across time in diachronic analysis. TSGNS is a practical and scalable method for embedding words such that the change of co-occurrence statistics of these words over time can be captured. It bypasses a major hurdle faced by previous methods and may help build a robust understanding of how vocabulary evolve with social and cultural change. Specifically, while enjoying the higher efficiency and better generalization of word embedding in low-dimensional dense vector space, the smooth alignment across time like in high-dimensional sparse vector space has been achieved naturally. The test results on a large corpus show the effectiveness of TSGNS in diachronic analysis and its advantage against current state-of-the-art, i.e., PPMI, SVD, SGNS and DW2V. Also, the scheme of TSGNS can also be applied to SVD.

\begin{acks}
Acknowledgement.
\end{acks}

\bibliographystyle{ACM-Reference-Format}
\bibliography{wsdm2020_xfx}


\begin{thebibliography}{30}


\ifx \showCODEN    \undefined \def \showCODEN     #1{\unskip}     \fi
\ifx \showDOI      \undefined \def \showDOI       #1{#1}\fi
\ifx \showISBNx    \undefined \def \showISBNx     #1{\unskip}     \fi
\ifx \showISBNxiii \undefined \def \showISBNxiii  #1{\unskip}     \fi
\ifx \showISSN     \undefined \def \showISSN      #1{\unskip}     \fi
\ifx \showLCCN     \undefined \def \showLCCN      #1{\unskip}     \fi
\ifx \shownote     \undefined \def \shownote      #1{#1}          \fi
\ifx \showarticletitle \undefined \def \showarticletitle #1{#1}   \fi
\ifx \showURL      \undefined \def \showURL       {\relax}        \fi
\providecommand\bibfield[2]{#2}
\providecommand\bibinfo[2]{#2}
\providecommand\natexlab[1]{#1}
\providecommand\showeprint[2][]{arXiv:#2}

\bibitem[\protect\citeauthoryear{Azarbonyad, Dehghani, Beelen, Arkut, Marx, and
  Kamps}{Azarbonyad et~al\mbox{.}}{2017}]%
        {Hosein:2017}
\bibfield{author}{\bibinfo{person}{Hosein Azarbonyad}, \bibinfo{person}{Mostafa
  Dehghani}, \bibinfo{person}{Kaspar Beelen}, \bibinfo{person}{Alexandra
  Arkut}, \bibinfo{person}{Maarten Marx}, {and} \bibinfo{person}{Jaap Kamps}.}
  \bibinfo{year}{2017}\natexlab{}.
\newblock \showarticletitle{Words are Malleable: Computing Semantic Shifts in
  Political and Media Discourse}.
\newblock \bibinfo{journal}{\emph{CoRR}}  \bibinfo{volume}{abs/1711.05603}
  (\bibinfo{year}{2017}).
\newblock
\showeprint[arxiv]{1711.05603}
\urldef\tempurl%
\url{http://arxiv.org/abs/1711.05603}
\showURL{%
\tempurl}


\bibitem[\protect\citeauthoryear{Bamler and Mandt}{Bamler and Mandt}{2013}]%
        {Robert:2017}
\bibfield{author}{\bibinfo{person}{Robert Bamler} {and}
  \bibinfo{person}{Stephan Mandt}.} \bibinfo{year}{2013}\natexlab{}.
\newblock \showarticletitle{Dynamic Word Embeddings}.
\newblock \bibinfo{journal}{\emph{arXiv:1702.08359v2}} (\bibinfo{year}{2013}).
\newblock


\bibitem[\protect\citeauthoryear{Barranco, Santos, and Hossain}{Barranco
  et~al\mbox{.}}{2018}]%
        {Roberto:2018}
\bibfield{author}{\bibinfo{person}{Roberto~Camacho Barranco},
  \bibinfo{person}{Raimundo F.~Dos Santos}, {and} \bibinfo{person}{M.~Shahriar
  Hossain}.} \bibinfo{year}{2018}\natexlab{}.
\newblock \showarticletitle{Tracking the Evolution of Words with
  Time-reflective Text Representations}.
\newblock \bibinfo{journal}{\emph{CoRR}}  \bibinfo{volume}{abs/1807.04441}
  (\bibinfo{year}{2018}).
\newblock
\showeprint[arxiv]{1807.04441}
\urldef\tempurl%
\url{http://arxiv.org/abs/1807.04441}
\showURL{%
\tempurl}


\bibitem[\protect\citeauthoryear{Bengio, Ducharme, Vincent, and Jauvin}{Bengio
  et~al\mbox{.}}{2003}]%
        {bengio2003neural}
\bibfield{author}{\bibinfo{person}{Yoshua Bengio}, \bibinfo{person}{R{\'e}jean
  Ducharme}, \bibinfo{person}{Pascal Vincent}, {and} \bibinfo{person}{Christian
  Jauvin}.} \bibinfo{year}{2003}\natexlab{}.
\newblock \showarticletitle{A neural probabilistic language model}.
\newblock \bibinfo{journal}{\emph{Journal of machine learning research}}
  \bibinfo{volume}{3}, \bibinfo{number}{Feb} (\bibinfo{year}{2003}),
  \bibinfo{pages}{1137--1155}.
\newblock


\bibitem[\protect\citeauthoryear{Bruni, Boleda, Baroni, and Tran}{Bruni
  et~al\mbox{.}}{2012}]%
        {bruni2012distributional}
\bibfield{author}{\bibinfo{person}{Elia Bruni}, \bibinfo{person}{Gemma Boleda},
  \bibinfo{person}{Marco Baroni}, {and} \bibinfo{person}{Nam-Khanh Tran}.}
  \bibinfo{year}{2012}\natexlab{}.
\newblock \showarticletitle{Distributional semantics in technicolor}. In
  \bibinfo{booktitle}{\emph{Proceedings of the 50th Annual Meeting of the
  Association for Computational Linguistics: Long Papers-Volume 1}}.
  Association for Computational Linguistics, \bibinfo{pages}{136--145}.
\newblock


\bibitem[\protect\citeauthoryear{Bullinaria and Levy}{Bullinaria and
  Levy}{2007}]%
        {bullinaria2007extracting}
\bibfield{author}{\bibinfo{person}{John~A Bullinaria} {and}
  \bibinfo{person}{Joseph~P Levy}.} \bibinfo{year}{2007}\natexlab{}.
\newblock \showarticletitle{Extracting semantic representations from word
  co-occurrence statistics: A computational study}.
\newblock \bibinfo{journal}{\emph{Behavior research methods}}
  \bibinfo{volume}{39}, \bibinfo{number}{3} (\bibinfo{year}{2007}),
  \bibinfo{pages}{510--526}.
\newblock


\bibitem[\protect\citeauthoryear{Devlin, Chang, Lee, and Toutanova}{Devlin
  et~al\mbox{.}}{2018}]%
        {devlin2018bert}
\bibfield{author}{\bibinfo{person}{Jacob Devlin}, \bibinfo{person}{Ming-Wei
  Chang}, \bibinfo{person}{Kenton Lee}, {and} \bibinfo{person}{Kristina
  Toutanova}.} \bibinfo{year}{2018}\natexlab{}.
\newblock \showarticletitle{Bert: Pre-training of deep bidirectional
  transformers for language understanding}.
\newblock \bibinfo{journal}{\emph{arXiv preprint arXiv:1810.04805}}
  (\bibinfo{year}{2018}).
\newblock


\bibitem[\protect\citeauthoryear{Garg, Schiebinger, Jurafsky, and Zou}{Garg
  et~al\mbox{.}}{2018}]%
        {garg2018word}
\bibfield{author}{\bibinfo{person}{Nikhil Garg}, \bibinfo{person}{Londa
  Schiebinger}, \bibinfo{person}{Dan Jurafsky}, {and} \bibinfo{person}{James
  Zou}.} \bibinfo{year}{2018}\natexlab{}.
\newblock \showarticletitle{Word embeddings quantify 100 years of gender and
  ethnic stereotypes}.
\newblock \bibinfo{journal}{\emph{Proceedings of the National Academy of
  Sciences}} \bibinfo{volume}{115}, \bibinfo{number}{16}
  (\bibinfo{year}{2018}), \bibinfo{pages}{E3635--E3644}.
\newblock


\bibitem[\protect\citeauthoryear{Gower and Dijksterhuis}{Gower and
  Dijksterhuis}{2004}]%
        {gower2004procrustes}
\bibfield{author}{\bibinfo{person}{John~C Gower} {and} \bibinfo{person}{Garmt~B
  Dijksterhuis}.} \bibinfo{year}{2004}\natexlab{}.
\newblock \bibinfo{booktitle}{\emph{Procrustes problems}}.
  Vol.~\bibinfo{volume}{30}.
\newblock \bibinfo{publisher}{Oxford University Press on Demand}.
\newblock


\bibitem[\protect\citeauthoryear{Grayson, Mulvany, Wade, Meaney, and
  Greene}{Grayson et~al\mbox{.}}{2017}]%
        {grayson2017exploring}
\bibfield{author}{\bibinfo{person}{Siobh{\'a}n Grayson}, \bibinfo{person}{Maria
  Mulvany}, \bibinfo{person}{Karen Wade}, \bibinfo{person}{Gerardine Meaney},
  {and} \bibinfo{person}{Derek Greene}.} \bibinfo{year}{2017}\natexlab{}.
\newblock \showarticletitle{Exploring the Role of Gender in 19th Century
  Fiction Through the Lens of Word Embeddings}. In
  \bibinfo{booktitle}{\emph{International Conference on Language, Data and
  Knowledge}}. Springer, \bibinfo{pages}{358--364}.
\newblock


\bibitem[\protect\citeauthoryear{Gulordava and Baroni}{Gulordava and
  Baroni}{2011}]%
        {gulordava2011distributional}
\bibfield{author}{\bibinfo{person}{Kristina Gulordava} {and}
  \bibinfo{person}{Marco Baroni}.} \bibinfo{year}{2011}\natexlab{}.
\newblock \showarticletitle{A distributional similarity approach to the
  detection of semantic change in the Google Books Ngram corpus}. In
  \bibinfo{booktitle}{\emph{Proceedings of the GEMS 2011 Workshop on
  GEometrical Models of Natural Language Semantics}}. Association for
  Computational Linguistics, \bibinfo{pages}{67--71}.
\newblock


\bibitem[\protect\citeauthoryear{Hamilton, Leskovec, and Jurafsky}{Hamilton
  et~al\mbox{.}}{2016a}]%
        {hamilton2016cultural}
\bibfield{author}{\bibinfo{person}{William~L Hamilton}, \bibinfo{person}{Jure
  Leskovec}, {and} \bibinfo{person}{Dan Jurafsky}.}
  \bibinfo{year}{2016}\natexlab{a}.
\newblock \showarticletitle{Cultural shift or linguistic drift? comparing two
  computational measures of semantic change}. In
  \bibinfo{booktitle}{\emph{Proceedings of the Conference on Empirical Methods
  in Natural Language Processing. Conference on Empirical Methods in Natural
  Language Processing}}, Vol.~\bibinfo{volume}{2016}. NIH Public Access,
  \bibinfo{pages}{2116}.
\newblock


\bibitem[\protect\citeauthoryear{Hamilton, Leskovec, and Jurafsky}{Hamilton
  et~al\mbox{.}}{2016b}]%
        {hamilton2016diachronic}
\bibfield{author}{\bibinfo{person}{William~L Hamilton}, \bibinfo{person}{Jure
  Leskovec}, {and} \bibinfo{person}{Dan Jurafsky}.}
  \bibinfo{year}{2016}\natexlab{b}.
\newblock \showarticletitle{Diachronic Word Embeddings Reveal Statistical Laws
  of Semantic Change}. In \bibinfo{booktitle}{\emph{Proceedings of the 54th
  Annual Meeting of the Association for Computational Linguistics (Volume 1:
  Long Papers)}}, Vol.~\bibinfo{volume}{1}. \bibinfo{pages}{1489--1501}.
\newblock


\bibitem[\protect\citeauthoryear{Harris}{Harris}{1954}]%
        {Harris:1954}
\bibfield{author}{\bibinfo{person}{Zellig~S. Harris}.}
  \bibinfo{year}{1954}\natexlab{}.
\newblock \showarticletitle{Distributional Structure}.
\newblock \bibinfo{journal}{\emph{Words}}  \bibinfo{volume}{10}
  (\bibinfo{year}{1954}), \bibinfo{pages}{146--162}.
\newblock


\bibitem[\protect\citeauthoryear{Jatowt and Duh}{Jatowt and Duh}{2014}]%
        {jatowt2014framework}
\bibfield{author}{\bibinfo{person}{Adam Jatowt} {and} \bibinfo{person}{Kevin
  Duh}.} \bibinfo{year}{2014}\natexlab{}.
\newblock \showarticletitle{A framework for analyzing semantic change of words
  across time}. In \bibinfo{booktitle}{\emph{Proceedings of the 14th
  ACM/IEEE-CS Joint Conference on Digital Libraries}}. IEEE Press,
  \bibinfo{pages}{229--238}.
\newblock


\bibitem[\protect\citeauthoryear{Kim, i~Chiu, Hanaki, Hegde, and Petrov}{Kim
  et~al\mbox{.}}{2014}]%
        {Kim:2014}
\bibfield{author}{\bibinfo{person}{Yoon Kim}, \bibinfo{person}{Yi i Chiu},
  \bibinfo{person}{Kentaro Hanaki}, \bibinfo{person}{Darshan Hegde}, {and}
  \bibinfo{person}{Slav Petrov}.} \bibinfo{year}{2014}\natexlab{}.
\newblock \bibinfo{title}{Temporal Analysis of Language through Neural Language
  Models}.
\newblock
\newblock


\bibitem[\protect\citeauthoryear{Kulkarni, Al-Rfou, Perozzi, and
  Skiena}{Kulkarni et~al\mbox{.}}{2015}]%
        {kulkarni2015statistically}
\bibfield{author}{\bibinfo{person}{Vivek Kulkarni}, \bibinfo{person}{Rami
  Al-Rfou}, \bibinfo{person}{Bryan Perozzi}, {and} \bibinfo{person}{Steven
  Skiena}.} \bibinfo{year}{2015}\natexlab{}.
\newblock \showarticletitle{Statistically significant detection of linguistic
  change}. In \bibinfo{booktitle}{\emph{Proceedings of the 24th International
  Conference on World Wide Web}}. International World Wide Web Conferences
  Steering Committee, \bibinfo{pages}{625--635}.
\newblock


\bibitem[\protect\citeauthoryear{Levy, Goldberg, and Dagan}{Levy
  et~al\mbox{.}}{2015}]%
        {levy2015improving}
\bibfield{author}{\bibinfo{person}{Omer Levy}, \bibinfo{person}{Yoav Goldberg},
  {and} \bibinfo{person}{Ido Dagan}.} \bibinfo{year}{2015}\natexlab{}.
\newblock \showarticletitle{Improving distributional similarity with lessons
  learned from word embeddings}.
\newblock \bibinfo{journal}{\emph{Transactions of the Association for
  Computational Linguistics}}  \bibinfo{volume}{3} (\bibinfo{year}{2015}),
  \bibinfo{pages}{211--225}.
\newblock


\bibitem[\protect\citeauthoryear{Mikolov, Chen, Corrado, and Dean}{Mikolov
  et~al\mbox{.}}{2013a}]%
        {mikolov2013efficient}
\bibfield{author}{\bibinfo{person}{Tomas Mikolov}, \bibinfo{person}{Kai Chen},
  \bibinfo{person}{Greg Corrado}, {and} \bibinfo{person}{Jeffrey Dean}.}
  \bibinfo{year}{2013}\natexlab{a}.
\newblock \showarticletitle{Efficient estimation of word representations in
  vector space}.
\newblock \bibinfo{journal}{\emph{arXiv preprint arXiv:1301.3781}}
  (\bibinfo{year}{2013}).
\newblock


\bibitem[\protect\citeauthoryear{Mikolov, Yih, and Zweig}{Mikolov
  et~al\mbox{.}}{2013b}]%
        {mikolov2013linguistic}
\bibfield{author}{\bibinfo{person}{Tomas Mikolov}, \bibinfo{person}{Wen-tau
  Yih}, {and} \bibinfo{person}{Geoffrey Zweig}.}
  \bibinfo{year}{2013}\natexlab{b}.
\newblock \showarticletitle{Linguistic regularities in continuous space word
  representations.}. In \bibinfo{booktitle}{\emph{hlt-Naacl}},
  Vol.~\bibinfo{volume}{13}. \bibinfo{pages}{746--751}.
\newblock


\bibitem[\protect\citeauthoryear{Peters, Neumann, Iyyer, Gardner, Clark, Lee,
  and Zettlemoyer}{Peters et~al\mbox{.}}{2018}]%
        {peters2018deep}
\bibfield{author}{\bibinfo{person}{Matthew Peters}, \bibinfo{person}{Mark
  Neumann}, \bibinfo{person}{Mohit Iyyer}, \bibinfo{person}{Matt Gardner},
  \bibinfo{person}{Christopher Clark}, \bibinfo{person}{Kenton Lee}, {and}
  \bibinfo{person}{Luke Zettlemoyer}.} \bibinfo{year}{2018}\natexlab{}.
\newblock \showarticletitle{Deep Contextualized Word Representations}. In
  \bibinfo{booktitle}{\emph{Proceedings of the 2018 Conference of the North
  American Chapter of the Association for Computational Linguistics: Human
  Language Technologies, Volume 1 (Long Papers)}}. \bibinfo{pages}{2227--2237}.
\newblock


\bibitem[\protect\citeauthoryear{Rong}{Rong}{2016}]%
        {Rong:2016}
\bibfield{author}{\bibinfo{person}{Xin Rong}.} \bibinfo{year}{2016}\natexlab{}.
\newblock \bibinfo{title}{word2vec Parameter Learning Explained}.
\newblock
\newblock


\bibitem[\protect\citeauthoryear{Rudolph and Blei}{Rudolph and Blei}{2018}]%
        {Rudolph:2018}
\bibfield{author}{\bibinfo{person}{Maja Rudolph} {and} \bibinfo{person}{David
  Blei}.} \bibinfo{year}{2018}\natexlab{}.
\newblock \showarticletitle{Dynamic Embeddings for Language Evolution}. In
  \bibinfo{booktitle}{\emph{Proceedings of the 2018 World Wide Web Conference}}
  \emph{(\bibinfo{series}{WWW '18})}. \bibinfo{publisher}{International World
  Wide Web Conferences Steering Committee}, \bibinfo{address}{Republic and
  Canton of Geneva, Switzerland}, \bibinfo{pages}{1003--1011}.
\newblock
\showISBNx{978-1-4503-5639-8}
\urldef\tempurl%
\url{https://doi.org/10.1145/3178876.3185999}
\showDOI{\tempurl}


\bibitem[\protect\citeauthoryear{Sagi, Kaufmann, and Clark}{Sagi
  et~al\mbox{.}}{2011}]%
        {sagi2011tracing}
\bibfield{author}{\bibinfo{person}{Eyal Sagi}, \bibinfo{person}{Stefan
  Kaufmann}, {and} \bibinfo{person}{Brady Clark}.}
  \bibinfo{year}{2011}\natexlab{}.
\newblock \showarticletitle{Tracing semantic change with latent semantic
  analysis}.
\newblock \bibinfo{journal}{\emph{Current methods in historical semantics}}
  (\bibinfo{year}{2011}), \bibinfo{pages}{161--183}.
\newblock


\bibitem[\protect\citeauthoryear{Turney and Pantel}{Turney and Pantel}{2010}]%
        {turney2010frequency}
\bibfield{author}{\bibinfo{person}{Peter~D Turney} {and}
  \bibinfo{person}{Patrick Pantel}.} \bibinfo{year}{2010}\natexlab{}.
\newblock \showarticletitle{From frequency to meaning: Vector space models of
  semantics}.
\newblock \bibinfo{journal}{\emph{Journal of artificial intelligence research}}
   \bibinfo{volume}{37} (\bibinfo{year}{2010}), \bibinfo{pages}{141--188}.
\newblock


\bibitem[\protect\citeauthoryear{Wijaya and Yeniterzi}{Wijaya and
  Yeniterzi}{2011}]%
        {wijaya2011understanding}
\bibfield{author}{\bibinfo{person}{Derry~Tanti Wijaya} {and}
  \bibinfo{person}{Reyyan Yeniterzi}.} \bibinfo{year}{2011}\natexlab{}.
\newblock \showarticletitle{Understanding semantic change of words over
  centuries}. In \bibinfo{booktitle}{\emph{Proceedings of the 2011
  international workshop on DETecting and Exploiting Cultural diversiTy on the
  social web}}. ACM, \bibinfo{pages}{35--40}.
\newblock


\bibitem[\protect\citeauthoryear{X. and G.}{X. and G.}{2016}]%
        {Liao:2016}
\bibfield{author}{\bibinfo{person}{Liao X.} {and} \bibinfo{person}{Cheng G.}}
  \bibinfo{year}{2016}\natexlab{}.
\newblock \showarticletitle{Analysing the Semantic Change Based on Word
  Embedding}.
\newblock \bibinfo{journal}{\emph{Natural Language Understanding and
  Intelligent Applications, LNCS}}  \bibinfo{volume}{10102}
  (\bibinfo{year}{2016}).
\newblock


\bibitem[\protect\citeauthoryear{Xu and Kemp}{Xu and Kemp}{2015}]%
        {xu2015computational}
\bibfield{author}{\bibinfo{person}{Yang Xu} {and} \bibinfo{person}{Charles
  Kemp}.} \bibinfo{year}{2015}\natexlab{}.
\newblock \showarticletitle{A Computational Evaluation of Two Laws of Semantic
  Change.}. In \bibinfo{booktitle}{\emph{CogSci}}.
\newblock


\bibitem[\protect\citeauthoryear{Yao, Sun, Ding, Rao, and Xiong}{Yao
  et~al\mbox{.}}{2018}]%
        {Yao:2018}
\bibfield{author}{\bibinfo{person}{Zijun Yao}, \bibinfo{person}{Yifan Sun},
  \bibinfo{person}{Weicong Ding}, \bibinfo{person}{Nikhil Rao}, {and}
  \bibinfo{person}{Hui Xiong}.} \bibinfo{year}{2018}\natexlab{}.
\newblock \showarticletitle{Dynamic Word Embeddings for Evolving Semantic
  Discovery}. In \bibinfo{booktitle}{\emph{Proceedings of the Eleventh ACM
  International Conference on Web Search and Data Mining}}
  \emph{(\bibinfo{series}{WSDM '18})}. \bibinfo{publisher}{ACM},
  \bibinfo{address}{New York, NY, USA}, \bibinfo{pages}{673--681}.
\newblock
\showISBNx{978-1-4503-5581-0}
\urldef\tempurl%
\url{https://doi.org/10.1145/3159652.3159703}
\showDOI{\tempurl}


\bibitem[\protect\citeauthoryear{Zhang, Jatowt, Bhowmick, and Tanaka}{Zhang
  et~al\mbox{.}}{2016}]%
        {7511732}
\bibfield{author}{\bibinfo{person}{Y. Zhang}, \bibinfo{person}{A. Jatowt},
  \bibinfo{person}{S.~S. Bhowmick}, {and} \bibinfo{person}{K. Tanaka}.}
  \bibinfo{year}{2016}\natexlab{}.
\newblock \showarticletitle{The Past is Not a Foreign Country: Detecting
  Semantically Similar Terms across Time}.
\newblock \bibinfo{journal}{\emph{IEEE Transactions on Knowledge and Data
  Engineering}} \bibinfo{volume}{28}, \bibinfo{number}{10} (\bibinfo{date}{Oct}
  \bibinfo{year}{2016}), \bibinfo{pages}{2793--2807}.
\newblock
\showISSN{1041-4347}
\urldef\tempurl%
\url{https://doi.org/10.1109/TKDE.2016.2591008}
\showDOI{\tempurl}


\end{thebibliography}


\end{document}